\PassOptionsToPackage{numbers,sort&compress}{natbib}
\documentclass{article}
\usepackage[dblblindworkshop, final]{neurips_2025}

\workshoptitle{New Perspectives in Advancing Graph Machine Learning}

\usepackage{graphicx}
\usepackage[utf8]{inputenc}
\usepackage[T1]{fontenc}   
\usepackage[pdfstartview={XYZ null null 1.2}]{hyperref}
\usepackage{url}           
\usepackage{booktabs}      
\usepackage{amsfonts}      
\usepackage{nicefrac}      
\usepackage{microtype}     
\usepackage{xcolor}       
\usepackage{amsmath}
\usepackage[none]{hyphenat}

\title{Learning Joint Embeddings of Function and Process Call Graphs for Malware Detection}

\author{%
    Kartikeya Aneja\thanks{Department of Electrical and Computer Engineering}\\
    University of Wisconsin-Madison\\
    Madison, WI, USA\\ 
    \texttt{kaneja@wisc.edu}\\ 
    \And
    Nagender Aneja\thanks{Bradley Department of Electrical and Computer Engineering}\\
    Virginia Tech\\
    Blacksburg, VA, USA\\
    \texttt{naneja@vt.edu}\\ 
    \And
    Murat Kantarcioglu\thanks{Department of Computer Science}\\
    Virginia Tech\\
    Blacksburg, VA, USA\\
    \texttt{muratk@vt.edu}
}

\begin{document}

\maketitle

\begin{abstract}
Software systems can be represented as graphs, capturing dependencies among functions and processes. An interesting aspect of software systems is that they can be represented as different types of graphs, depending on the extraction goals and priorities. For example, function calls within the software can be captured to create function call graphs, which highlight the relationships between functions and their dependencies. Alternatively, the processes spawned by the software can be modeled to generate process interaction graphs, which focus on runtime behavior and inter-process communication. While these graph representations are related, each captures a distinct perspective of the system, providing complementary insights into its structure and operation. While previous studies have leveraged graph neural networks (GNNs) to analyze software behaviors, most of this work has focused on a single type of graph representation. The joint modeling of both function call graphs and process interaction graphs remains largely underexplored, leaving opportunities for deeper, multi-perspective analysis of software systems. This paper presents a pipeline for constructing and training \textit{Function Call Graphs (FCGs)} and \textit{Process Call Graphs (PCGs)} and learning joint embeddings. We demonstrate that joint embeddings outperform a single-graph model. In this paper, we propose GeminiNet, a unified neural network approach that learns joint embeddings from both FCGs and PCGs. We construct a new dataset of 635 Windows executables (318 malicious and 317 benign), extracting FCGs via Ghidra and PCGs via Any.Run sandbox. GeminiNet employs dual graph convolutional branches with an adaptive gating mechanism that balances contributions from static and dynamic views. Experiments with five-fold cross-validation demonstrate that GeminiNet achieves a mean F1 score of $0.85$ with a standard deviation of $ 0.06$-$0.09$ for SGC and GCN, and a maximum F1 score of up to $0.94$ for the best configuration. Extensive experiments with multiple graph neural networks demonstrate that GeminiNet's joint embeddings consistently outperform single modalities and the merged-graph modality. 
\end{abstract}

\section{Introduction}
Graph representations of software have gained importance in malware analysis and vulnerability detection. Different graph modalities capture different software perspectives. Function Call Graphs (FCGs) capture intra-program dependencies, showing how various functions call each other, while Process Call Graphs (PCGs) capture runtime execution behavior, illustrating how various processes interact with the software.

Previous work generally extracted and analyzed FCGs and PCGs separately. For example,
\citet{freitas2021large, chen2023guided, ling2022malgraph, gulmez2021graph, kakisim2020metamorphic, niu2020opcode, feng2021android} use function call graphs; \citet{gao2021gdroid, hou2017hindroid, pei2020amalnet, gu2024gsedroid} use API call graphs; \citet{busch2021nf} use network flow graph; \citet{peng2025evading, yan2019classifying} use control flow graphs.

Building on these insights, we propose GeminiNet, a unified framework that jointly learns from both FCGs and PCGs to overcome the limitations of analyzing them separately. Specifically, FCGs capture static control-flow structure, while PCGs reflect dynamic execution traces and inter-process interactions. Relying on only one modality risks adversarial fragility, whereas combining both provides a more resilient representation. GeminiNet addresses this by learning joint embeddings across the two graph modalities, thereby improving the robustness of malware detection. Beyond malware detection, this approach can be generalized to software vulnerability analysis or binary similarity detection.
Our key contributions include:
\begin{itemize}
    \item A dataset construction pipeline for FCGs and PCGs from Windows executables
    \item GeminiNet, which is a Graph Convolutional Neural Network with a joint embedding approach that fuses representations across FCGs and PCGs
    \item Adaptive gating mechanism to fuse joint embeddings
    \item Joint node features integrating Local Degree Profile with the file-level Shanon Entropy
\end{itemize}

\begin{figure}[htbp!]
    \centering
    \includegraphics[width=1.0\linewidth]{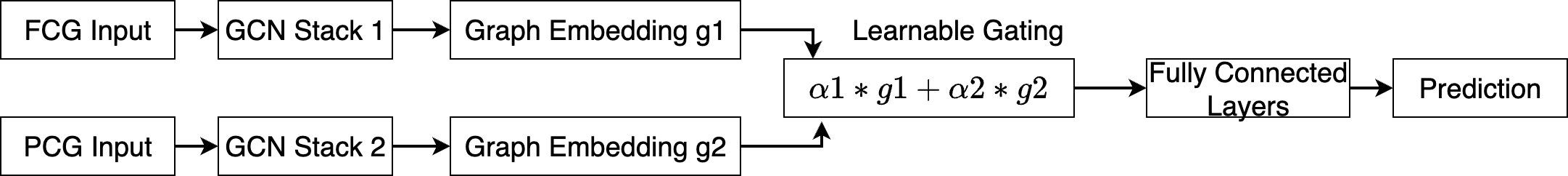}
    \caption{Architecture of GeminiNet}
    \label{fig:GeminiNet}
\end{figure}

\section{Methodology}
We used Windows Executable files to create FCGs and PCGs. Each node in FCG represents a function, while in PCG it represents a process. Edges represent call or communication relationships between functions or processes corresponding to software behavior.

\subsection{Dataset Construction}
This section outlines the steps and tools for creating FCGs and PCGs from Windows executable files. The dataset comprises 635 Windows Portable Executable (PE) files, consisting of 318 malicious and 317 benign samples.

\subsubsection{Function Call Graphs (FCGs)}
The FCGs are constructed from Windows Executables using Ghidra \cite{NSA_2025_Ghidra}. Ghidra is a software reverse engineering framework created and maintained by the National Security Agency Research Directorate. Each executable was decompiled, and the resulting FCG was created with nodes representing functions and directed edges representing invocations from one function to another. To enable the graph neural network, each function name and call target is replaced with a numeric identifier. Across all $635$ executables, the FCGs contained a total of $449,960$ nodes and $1,048,741$ edges. 

\subsubsection{Process Call Graphs (FCGs)}
Dynamic process-level behavior is captured using \textit{Any.Run} malware sandbox. Each executable was executed in a controlled environment for 60 seconds. \citet{kuchler2021does} conducted a study on execution time on malware behavior. The authors observed that most samples execute for under two minutes, achieving 98\% code coverage. During execution, all process creation and inter-process interactions are logged to create PCG. In the PCGs, the nodes represent spawned processes, while directed edges denote communication or creation relationships between processes. Across all samples, the PCGs consisted of $3,053$ nodes and $2,663$ edges, which is significantly smaller than the FCGs since the number of processes spawned is less than the number of functions. Both FCGs and PCGs complement each other, as shown in the results.

\subsubsection{Combined Graph Dataset}
For experiments involving both FCGs (static) and PCGs (dynamic), we aligned FCGs and PCGs per binary. This resulted in a joint dataset of 635 graph pairs, comprising a total of $453,013$ nodes and $1,051,404$ edges. The combined dataset can support a broader range of experimentation. Since, each executable is represented by both FCG and PCG, we can compare static-only, dynamic-only, and how fusion can improve the robustness. The paired dataset also enabled edge-level merging and dual-branch embedding fusion. This advances multimodal graph learning in software security. Thus, the combined dataset supports experiments in four main configurations (i) Single-graph FCG, (ii) Single-graph PCG, (iii) Merged Graph (edges of both graphs combined into a single graph, but these two graphs are not connected naturally), and (iv) Dual-branch graph (separate convolutional encoders for FCG and PCG embeddings fused via GemniNet).

\subsection{Graph Node Features}
To prepare the graphs for GNN, the initial node features are constructed from (i) Local Degree Profile (LDP): captures degree-based statistics of each node's neighborhood, (ii) Shanon Entropy: file-level feature measuring randomness in the binary. The computed entropy value is assigned uniformly to all nodes in the graph, and (iii) LDP+Entropy: concatenate both features: LDP features (structural) and Shanon Entropy features (global statistical) information per node. 

The LDP is a structural graph feature that characterizes each node based on its degree and the degree statistics of its local neighborhood. Specifically, the LDP captures the degree of the node itself, as well as the minimum, maximum, mean, and standard deviation of the degrees of its neighbors. Thus, each node has five features providing information about the local topology around it. 

The entropy is derived using Shannon entropy to capture the overall information content of each sample. The entropy is computed directly from the raw binary representation of the input file by treating each byte as a symbol and applying Shannon's definition of entropy $H(X) = - \sum\limits_{i} p_i \log_2 p_i$, where $p_i$ is the probability of symbol i. This value indicates the degree of randomness in the file. Highly redundant files yield lower entropy, as expected in benign executables, while more diverse content produces higher entropy, as expected in malware. Since the entropy is a global property of the file rather than the node, the same entropy was assigned to all nodes as an initial feature. Thus, there is one feature per node providing information about the randomness of the binary.

The combined feature, LDP+Entropy, combines both features from LDP and Shannon entropy. Thus, each node's feature vector integrates local structural information with the global characteristic of the sample. This fusion provides a richer and more discriminative representation, which improved the predictive performance in our experiments. Thus, in this case, we have six features per node providing local and global information.

We experimented with LDP, Entropy, and LDP+Entropy using a single FCG graph, a single PCG Graph, merging both graphs by combining the edges of both graphs while maintaining one convolutional graph setting, and dual convolutional stacks, whose embeddings are fused using a weighted sum. Our results show that the fusion improves the predictive power. 

\begin{table}[bhtp!]
\caption{Validation F1 scores for top configurations (5-fold CV)}
\label{tab:f1top}
\resizebox{\textwidth}{!}{%
\begin{tabular}{|r|r|r|r|r|r|r|r|r|r|r|r|r|}
\hline
graph\_type & feature & model\_arch & join\_embeddings & layer & fc & dim & scheduler & mean & std & min  & median  & max \\ \hline

both graphs & ldp+entropy & SGC       & wsum & 6 & 6 & 64 & ReduceLROnPlateau & 0.85 & 0.09 & 0.69 & 0.88 & 0.91 \\ \hline

both graphs & ldp+entropy & SGC       & wsum & 6 & 6 & 32 & OneCycleLR        & 0.85 & 0.06 & 0.78 & 0.83 & 0.94 \\ \hline

both graphs & ldp+entropy & GCN       & wsum & 6 & 6 & 64 & OneCycleLR        & 0.85 & 0.06 & 0.76 & 0.84 & 0.94 \\ \hline

both graphs & ldp+entropy & GCN       & wsum & 6 & 3 & 64 & OneCycleLR        & 0.84 & 0.06 & 0.79 & 0.85 & 0.92 \\ \hline

both graphs & ldp         & GCN       & wsum & 6 & 6 & 32 & OneCycleLR        & 0.84 & 0.05 & 0.76 & 0.85 & 0.89 \\ \hline
\end{tabular}%
}
\end{table}

\begin{table}[bhtp!]
\caption{Validation F1 scores for merged graphs (5-fold CV)}
\label{tab:f1top_merged}
\resizebox{\textwidth}{!}{%
\begin{tabular}{|r|r|r|r|r|r|r|r|r|r|r|}
\hline
feature     & model\_arch & layer & fc & dim & scheduler         & mean & std  & min  & median & max  \\ \hline

ldp+entropy & SGC         & 4     & 2  & 64  & OneCycleLR        & 0.73 & 0.03 & 0.70 & 0.72 & 0.77 \\ \hline

ldp+entropy & SGC         & 5     & 2  & 32  & OneCycleLR        & 0.72 & 0.05 & 0.68 & 0.71 & 0.78  \\ \hline

ldp         & SGC         & 4     & 2  & 32  & OneCycleLR        & 0.72 & 0.04 & 0.66 & 0.72  & 0.78     \\ \hline

ldp         & SGC         & 4     & 2  & 64  & ReduceLROnPlateau & 0.72 & 0.01 & 0.71 & 0.72  & 0.73    \\ \hline

ldp+entropy & SGC         & 4     & 2  & 32  & OneCycleLR        & 0.72 & 0.06 & 0.63 & 0.72  & 0.80    \\ \hline
\end{tabular}%
}
\end{table}

\begin{table}[bhtp!]
\caption{Validation F1 scores for single-graph Function Call Graph (FCG) configurations (5-fold CV)}
\label{tab:f1top_fcg}
\resizebox{\textwidth}{!}{%
\begin{tabular}{|r|r|r|r|r|r|r|r|r|r|r|}
\hline
feature     & model\_arch & layer & fc & dim & scheduler         & mean & std  & min   & median & max \\ \hline

ldp+entropy & SGC         & 5     & 2  & 64  & ReduceLROnPlateau & 0.72 & 0.04 & 0.67 & 0.71 & 0.77 \\ \hline

ldp+entropy & SGC         & 5     & 2  & 32  & ReduceLROnPlateau & 0.70 & 0.05 & 0.64 & 0.72 & 0.76  \\ \hline

ldp+entropy & SGC         & 4     & 2  & 64  & OneCycleLR        & 0.70 & 0.02 & 0.68 & 0.70 & 0.72 \\ \hline

ldp+entropy & SGC         & 4     & 2  & 64  & ReduceLROnPlateau & 0.69 & 0.03 & 0.65 & 0.69  & 0.72 \\ \hline

ldp+entropy & SGC         & 5     & 2  & 64  & OneCycleLR        & 0.68 & 0.03 & 0.65 & 0.68 & 0.71 \\ \hline
\end{tabular}%
}
\end{table}

\begin{table}[bhtp!]
\caption{Validation F1 scores for single-graph Process Call Graph (PCG) configurations (5-fold CV)}
\label{tab:f1top_pcg}
\resizebox{\textwidth}{!}{%
\begin{tabular}{|r|r|r|r|r|r|r|r|r|r|r|}
\hline
feature     & model\_arch & layer & fc & dim & scheduler  & mean & std  & min & median & max  \\ \hline

ldp+entropy & GIN         & 5     & 2  & 64  & OneCycleLR & 0.83 & 0.04 & 0.76 & 0.83 & 0.87 \\ \hline

ldp+entropy & GIN         & 4     & 2  & 32  & OneCycleLR & 0.82 & 0.03 & 0.78 & 0.82  & 0.86 \\ \hline

ldp+entropy & GCN         & 5     & 2  & 64  & OneCycleLR & 0.82 & 0.03 & 0.78 &  0.82   &  0.85 \\ \hline

ldp+entropy & GIN         & 5     & 2  & 32  & OneCycleLR & 0.82 & 0.04 & 0.78 & 0.83 & 0.87 \\ \hline

ldp+entropy & SGC         & 4     & 2  & 32  & OneCycleLR & 0.81 & 0.02 & 0.78 & 0.82 &  0.83 \\ \hline
\end{tabular}%
}
\end{table}

\subsection{GeminiNet}
We propose GeminiNet, a unified neural architecture that jointly learns from the FCGs and PCGs. The core idea is to treat static and dynamic software views as two complementary modalities and fuse their embeddings into a shared embedding space. Figure \ref{fig:GeminiNet} shows an architecture consisting of two parallel GCN branches. One branch encodes FCG, and the other branch encodes PCG. Each branch produces graph-level embeddings, which are then fused through a learnable gating mechanism that adaptively weights static and dynamic information. The joint embedding is then passed through fully connected layers for malware classification. GeminiNet generalizes to the single-graph setting by turning off the dual-branch architecture. In this case, only one GCN branch becomes active, and the gating mechanism is bypassed.

Formally, let $G_1 = (V_1, E_1)$ denote the FCG and $G_2 = (V_2, E_2)$ denote the PCG. Each graph $G_n$ is processed by a branch specific GCN stack $H_m = GCN_n(X_n, E_n),~n\in\{1,2\}$, where $X_n$ are node features. A global pooling operator aggregates node embeddings into a graph-level representation $g_n$. 

A key innovation feature of GeminiNet is the learnable gating mechanism that adaptively balances the contribution of FCG and PCG embeddings. Rather than statically concatenating or averaging, we introduce a trainable gate vector $\alpha = \text{softmax(w)}$, where $w$ is a learnable parameter. The final joint embedding is given by $ g = \alpha_1 g_1 + \alpha_2 g_2$, where $\alpha_1 + \alpha_2 = 1$ and $\alpha_i \ge 0$. This combination allows the model to emphasize the most informative modality, while preserving contributions from both.

The joint embedding $g$ is passed through fully connected layers with ReLU activation and dropout regularization. The final output is a probability distribution over malware and benign classes, $ \hat{y} = \text{softmax}(MLP(g))$. We conducted experiments using multiple convolutional and fully connected layers.

\section{Results}
We primarily evaluated GeminiNet using a GCN backbone. Additionally, we tested alternative architectures, including the Graph Isomorphism Network (GIN) \cite{xu2018powerful}, Graph Sample and Aggregate (GraphSAGE), and Simplifying Graph Convolutional Networks (SGC) \cite{wu2019simplifying}, as well as the base model, Multi Layer Perceptron. All models were trained using five-fold cross-validation across multiple architectural and training configurations. 

Table \ref{tab:f1top} reports validation performance across the top configurations, with the following columns:
\begin{itemize}
    \item graph\_type: whether the model used a single graph (FCG or PCG) or both graphs together
    \item feature: initial node feature representation, LDP, Entropy, or LDP+Entropy
    \item model\_arch: GNN architecturee.g., SGC, GCN, GraphSAGE, GIN, MLP
    \item join\_embeddings: strategy for combining embeddings weighted sum (wsum) or edges merged
    \item layer: number of graph convolutional layers
    \item fc: number of fully connected layers following the graph convolutional stack
    \item dim: hidden dimension size of the embeddings
    \item scheduler: learning rate scheduling strategy (e.g., OneCycleLR, ReduceLROnPlateau)
    \item mean: mean F1 score across five folds
    \item std: standard deviation of F1 scores across five folds
    \item min/median/max: minimum, median, and maximum of F1 scores across folds
\end{itemize}

\subsection{Ablation study (k-fold)}
We performed a controlled ablation study using 5-fold cross-validation, holding folds fixed across multiple GCN variants, namely GIN, GraphSAGE, SGC, and MLP, in addition to GCN-based GeminiNet. The GIN, GraphSAGE, SGC, and MLP were all modified to consider both PCGs and FCGs as similar to GeminiNet.

We conducted multiple experiments, which can be categorized into two main categories: single graph networks and dual graph networks. In the single graph networks, we used function call graphs and process call graphs with one graph at a time. In the dual graphs, we first combined the edge lists of both graphs by renumbering the nodes so that they form a single graph suitable for neural networks. In the second approach, we processed both graphs so the convolutional weights are not shared, and the embeddings are joined with a weighted sum, with weights being trained. In all of these experiments, we also used different starting features of nodes: LDP, Entropy, and LDP+Entropy. 

Table \ref{tab:f1top} provides the top configurations for validation F1 score. The strongest configurations achieved a mean in the range of \{0.84 - 085\}. With both graphs input, an initial feature as the LDP+Entropy feature combination and a weighted sum (wsum) as joint embeddings, SGC and GCN consistently perform better. Other families, GraphSAGE and GIN, also performed well. Across the top configuration, the feature combination LDP+Entropy and joint embeddings, presented as a weighted sum, appeared universally, suggesting that the design choice contributes more to performance than the architecture alone.

Table \ref{tab:f1top_merged} provides the top configurations for variation of F1 score for the merged graph, where the FCG and PCG edge lists were combined into a single unified graph before learning. Compared to the dual branch GemniNet approach, the merged setting achieved noticeably lower performance, with mean F1 scores clustering \{0.71--0.73\}. However, the top configuration suggests LDP+Entropy as the best feature combination.

For completeness, we also evaluated single-graph models on FCGs and PCGs independently. Table \ref{tab:f1top_fcg} summarizes the results for FCG-based models, where performance was modest, with the best mean F1 score of 0.72 obtained by SGC with LDP+Entropy features. Table \ref{tab:f1top_pcg} shows that PCG-based models were stronger than FCG, achieving mean F1 scores above 0.80.

\begin{table}[t]
\centering
\caption{Kruskal-Wallis test and pairwise p-values.}
\label{tab:p_both_single}
\begin{tabular}{|l|c|c|c|c|}
\hline
{} & both\_merged & both\_wsum & single\_fcg & single\_pcg \\
\hline
Kruskal-Wallis $p$ & \multicolumn{4}{c|}{3.86e-76} \\
\hline
both\_merged & 1.00e+00 & 2.15e-34 & 2.31e-06 & 2.41e-11 \\
\hline
both\_wsum   & 2.15e-34 & 1.00e+00 & 2.04e-67 & 3.23e-07 \\
\hline
single\_fcg  & 2.31e-06 & 2.04e-67 & 1.00e+00 & 1.83e-32 \\
\hline
single\_pcg  & 2.41e-11 & 3.23e-07 & 1.83e-32 & 1.00e+00 \\
\hline
\end{tabular}
\end{table}

\begin{figure*}[t]
    \centering
    \begin{minipage}{0.48\linewidth}
        \centering
        \includegraphics[width=\linewidth]{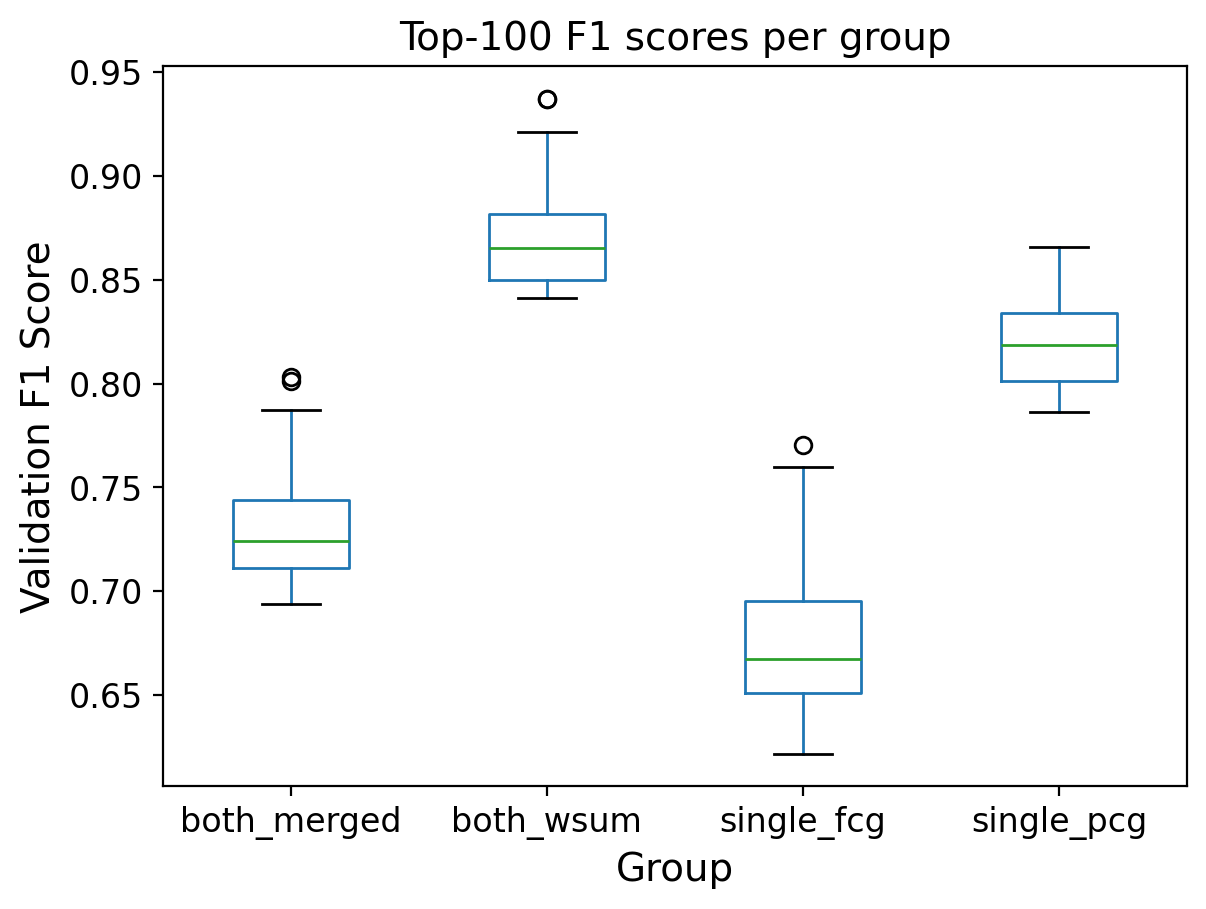}
        \caption{Variability and median performance of Top-100 validation F1 scores across groups}
        \label{fig:f1_100_both_single}
    \end{minipage}\hfill
    \begin{minipage}{0.48\linewidth}
        \centering
        \includegraphics[width=\linewidth]{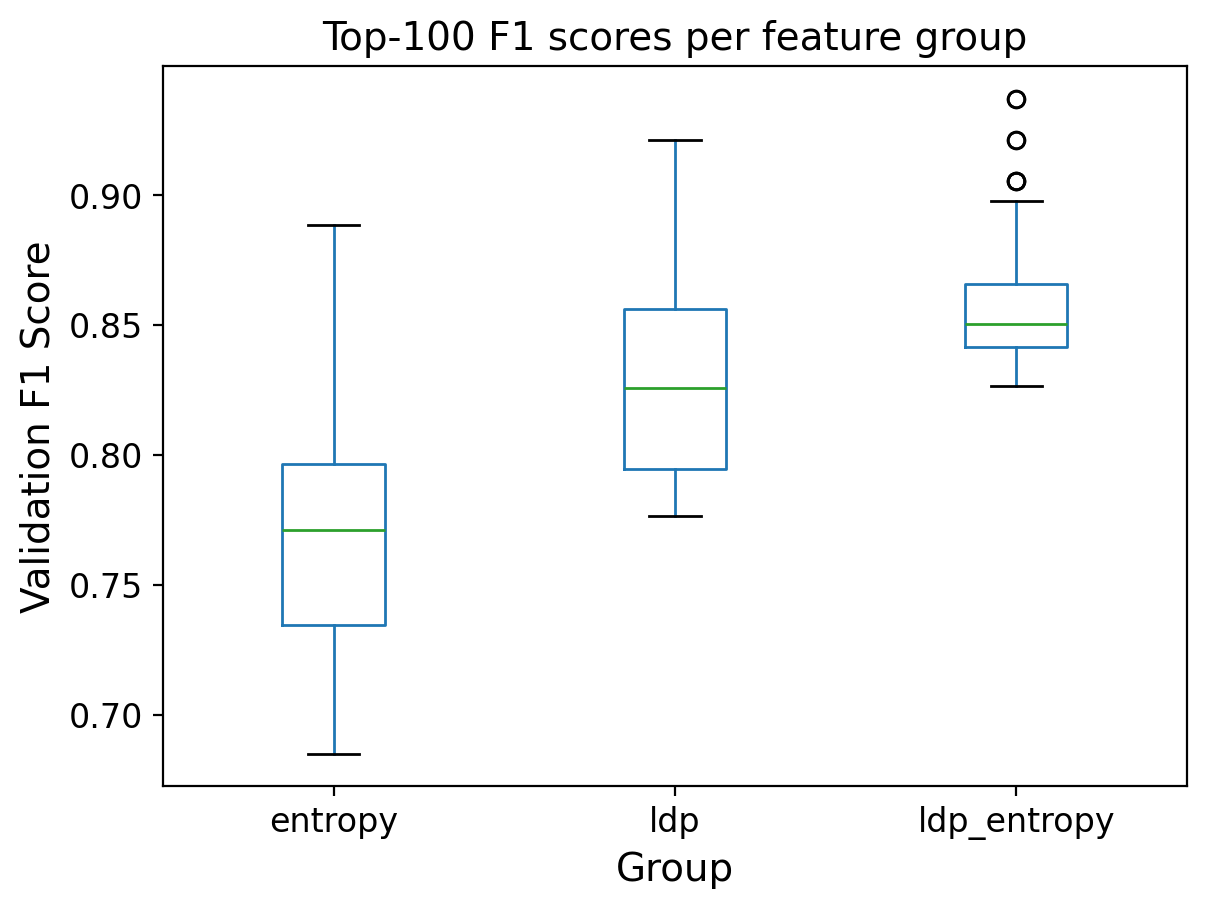}
        \caption{Variability and median performance of Top-100 validation F1 scores across groups}
        \label{fig:f1_100_feature}
    \end{minipage}
\end{figure*}

We compared four graph-based configurations: both merged, both weighted sum, single FCG, and single PCG, using the top 100 validation scores per group, regardless of model architecture, number of layers, number of neurons per hidden layer, or scheduler used. Table \ref{tab:p_both_single} shows that the Kruskal–Wallis test indicated highly significant differences across groups $ p=3.86\times 10^{-76}$. The Post-hoc Dunn test confirmed that all pairwise comparisons are statistically significant after Bonferroni correction. As shown in Figure \ref{fig:f1_100_both_single}, the both\_wsum configuration achieved the highest and most consistent performance (median $ F1 \approx 0.87 $), followed by single\_pcg (median $\approx 0.82 $). In contrast, both\_merged (median $\approx 0.72$) and single\_fcg (median $\approx 0.67$) performed significantly worse. These results highlight the advantage of weighted-sum embedding over merged representations and suggest that process call graphs yield stronger predictive performance than function call graphs when used individually.

\begin{table}[htbp!]
\centering
\caption{Kruskal-Wallis test result and pairwise p-values}
\label{p_feature_ldp_entropy}
\begin{tabular}{lccc}
\hline
{} & means\_entropy & means\_ldp & means\_ldp\_entropy \\
\hline
Kruskal-Wallis $p$ & \multicolumn{3}{c}{2.57e-33} \\
\hline
means\_entropy     & 1.00e+00 & 3.20e-12 & 1.01e-33 \\
means\_ldp         & 3.20e-12 & 1.00e+00 & 1.18e-06 \\
means\_ldp\_entropy& 1.01e-33 & 1.18e-06 & 1.00e+00 \\
\hline
\end{tabular}
\end{table}

We also conducted a comparative evaluation of three feature configurations: entropy, LDP, and their combination (LDP+entropy) using the top 100 validation F1 scores from each group. As shown in Figure \ref{p_feature_ldp_entropy}, a Kruskal–Wallis test revealed highly significant differences across the groups ($p \approx 2.57 × 10^{-33}$). Post-hoc Dunn tests with Bonferroni correction confirmed that all pairwise comparisons were statistically significant: ldp+entropy outperformed entropy ($p \approx 1.01 \times 10^{33}$), and ldp+entropy also outperformed ldp ($p \approx 1.18 \times 10^{-6}$). ldp outperformed entropy ($p \approx 3.20 \times 10^{-12}$), 

As illustrated in Figure \ref{fig:f1_100_feature}, the ldp+entropy group achieved the highest and most consistent F1 scores (median $\approx 0.85$), followed by ldp (median $\approx 0.82$), while entropy yielded the weakest results (median $\approx 0.77$). These findings demonstrate that combining entropy with LDP provides advantage, producing significantly stronger and more stable model performance than either feature alone.

\section{Conclusion}
In this work, we introduce GeminiNet, a unified graph neural architecture that combines static FCGs and dynamic PCGs for malware detection. We construct a new dataset of 635 Windows executables and design dual-branched GCN encoders with an adaptive gating mechanism. We demonstrate that fusing static and dynamic graphs provides more robust embeddings than either modality alone. The results highlight three key insights: (i) combining structural and behavior graphs mitigates the weakness of purely static or dynamic analysis, (ii) joint node features integrating Local Degree Profile and Entropy provide a more discriminative representation, and (iii) adaptive modality weighting yields flexibility to prioritize the most informative graph type per sample. Future work will explore scaling to larger datasets and investigating explainability techniques.

\bibliographystyle{unsrtnat}
\bibliography{refs}

\begin{thebibliography}{18}
\providecommand{\natexlab}[1]{#1}
\providecommand{\url}[1]{\texttt{#1}}
\expandafter\ifx\csname urlstyle\endcsname\relax
  \providecommand{\doi}[1]{doi: #1}\else
  \providecommand{\doi}{doi: \begingroup \urlstyle{rm}\Url}\fi

\bibitem[Freitas and Dong(2021)]{freitas2021large}
Scott Freitas and Yuxiao Dong.
\newblock A large-scale database for graph representation learning.
\newblock \emph{Advances in neural information processing systems}, 2021.

\bibitem[Chen et~al.(2023)Chen, Lin, Huang, Lei, and Huang]{chen2023guided}
Yi-Hsien Chen, Si-Chen Lin, Szu-Chun Huang, Chin-Laung Lei, and Chun-Ying
  Huang.
\newblock Guided malware sample analysis based on graph neural networks.
\newblock \emph{IEEE Transactions on Information Forensics and Security},
  18:\penalty0 4128--4143, 2023.

\bibitem[Ling et~al.(2022)Ling, Wu, Deng, Qu, Zhang, Zhang, Ma, Wang, Wu, and
  Ji]{ling2022malgraph}
Xiang Ling, Lingfei Wu, Wei Deng, Zhenqing Qu, Jiangyu Zhang, Sheng Zhang,
  Tengfei Ma, Bin Wang, Chunming Wu, and Shouling Ji.
\newblock Malgraph: Hierarchical graph neural networks for robust windows
  malware detection.
\newblock In \emph{IEEE INFOCOM 2022-IEEE Conference on Computer
  Communications}, pages 1998--2007. IEEE, 2022.

\bibitem[G{\"u}lmez and Sogukpinar(2021)]{gulmez2021graph}
Sibel G{\"u}lmez and Ibrahim Sogukpinar.
\newblock Graph-based malware detection using opcode sequences.
\newblock In \emph{2021 9th International Symposium on Digital Forensics and
  Security (ISDFS)}, pages 1--5. IEEE, 2021.

\bibitem[Kakisim et~al.(2020)Kakisim, Nar, and
  Sogukpinar]{kakisim2020metamorphic}
Arzu~Gorgulu Kakisim, Mert Nar, and Ibrahim Sogukpinar.
\newblock Metamorphic malware identification using engine-specific patterns
  based on co-opcode graphs.
\newblock \emph{Computer Standards \& Interfaces}, 71:\penalty0 103443, 2020.

\bibitem[Niu et~al.(2020)Niu, Cao, Zhang, Ding, Zhang, and Li]{niu2020opcode}
Weina Niu, Rong Cao, Xiaosong Zhang, Kangyi Ding, Kaimeng Zhang, and Ting Li.
\newblock Opcode-level function call graph based android malware classification
  using deep learning.
\newblock \emph{Sensors}, 20\penalty0 (13):\penalty0 3645, 2020.

\bibitem[Feng et~al.(2021)Feng, Ma, Li, Ma, Xi, and Lu]{feng2021android}
Pengbin Feng, Jianfeng Ma, Teng Li, Xindi Ma, Ning Xi, and Di~Lu.
\newblock Android malware detection via graph representation learning.
\newblock \emph{Mobile Information Systems}, 2021\penalty0 (1):\penalty0
  5538841, 2021.

\bibitem[Gao et~al.(2021)Gao, Cheng, and Zhang]{gao2021gdroid}
Han Gao, Shaoyin Cheng, and Weiming Zhang.
\newblock Gdroid: Android malware detection and classification with graph
  convolutional network.
\newblock \emph{Computers \& Security}, 106:\penalty0 102264, 2021.

\bibitem[Hou et~al.(2017)Hou, Ye, Song, and Abdulhayoglu]{hou2017hindroid}
Shifu Hou, Yanfang Ye, Yangqiu Song, and Melih Abdulhayoglu.
\newblock Hindroid: An intelligent android malware detection system based on
  structured heterogeneous information network.
\newblock In \emph{Proceedings of the 23rd ACM SIGKDD international conference
  on knowledge discovery and data mining}, pages 1507--1515, 2017.

\bibitem[Pei et~al.(2020)Pei, Yu, and Tian]{pei2020amalnet}
Xinjun Pei, Long Yu, and Shengwei Tian.
\newblock Amalnet: A deep learning framework based on graph convolutional
  networks for malware detection.
\newblock \emph{Computers \& Security}, 93:\penalty0 101792, 2020.

\bibitem[Gu et~al.(2024)Gu, Zhu, Han, Li, and Zhao]{gu2024gsedroid}
Jintao Gu, Hongliang Zhu, Zewei Han, Xiangyu Li, and Jianjin Zhao.
\newblock Gsedroid: Gnn-based android malware detection framework using
  lightweight semantic embedding.
\newblock \emph{Computers \& Security}, 140:\penalty0 103807, 2024.

\bibitem[Busch et~al.(2021)Busch, Kocheturov, Tresp, and Seidl]{busch2021nf}
Julian Busch, Anton Kocheturov, Volker Tresp, and Thomas Seidl.
\newblock Nf-gnn: network flow graph neural networks for malware detection and
  classification.
\newblock In \emph{Proceedings of the 33rd international conference on
  scientific and statistical database management}, pages 121--132, 2021.

\bibitem[Peng et~al.(2025)Peng, Yu, Zhao, Ding, Yang, Zhang, Han, Zhang, Ji,
  and Zhong]{peng2025evading}
Hao Peng, Zehao Yu, Dandan Zhao, Zhiguo Ding, Jieshuai Yang, Bo~Zhang, Jianming
  Han, Xuhong Zhang, Shouling Ji, and Ming Zhong.
\newblock Evading control flow graph based gnn malware detectors via active
  opcode insertion method with maliciousness preserving.
\newblock \emph{Scientific Reports}, 15\penalty0 (1):\penalty0 9174, 2025.

\bibitem[Yan et~al.(2019)Yan, Yan, and Jin]{yan2019classifying}
Jiaqi Yan, Guanhua Yan, and Dong Jin.
\newblock Classifying malware represented as control flow graphs using deep
  graph convolutional neural network.
\newblock In \emph{2019 49th annual IEEE/IFIP international conference on
  dependable systems and networks (DSN)}, pages 52--63. IEEE, 2019.

\bibitem[{National Security Agency}(2025)]{NSA_2025_Ghidra}
{National Security Agency}.
\newblock Ghidra – software reverse engineering framework.
\newblock \url{https://github.com/NationalSecurityAgency/ghidra}, 2025.
\newblock Accessed: 2025-06-05.

\bibitem[K{\"u}chler et~al.(2021)K{\"u}chler, Mantovani, Han, Bilge, and
  Balzarotti]{kuchler2021does}
Alexander K{\"u}chler, Alessandro Mantovani, Yufei Han, Leyla Bilge, and Davide
  Balzarotti.
\newblock Does every second count? time-based evolution of malware behavior in
  sandboxes.
\newblock In \emph{NDSS 2021, Network and Distributed Systems Security
  Symposium}. Internet Society, 2021.

\bibitem[Xu et~al.(2018)Xu, Hu, Leskovec, and Jegelka]{xu2018powerful}
Keyulu Xu, Weihua Hu, Jure Leskovec, and Stefanie Jegelka.
\newblock How powerful are graph neural networks?
\newblock \emph{arXiv preprint arXiv:1810.00826}, 2018.

\bibitem[Wu et~al.(2019)Wu, Souza, Zhang, Fifty, Yu, and
  Weinberger]{wu2019simplifying}
Felix Wu, Amauri Souza, Tianyi Zhang, Christopher Fifty, Tao Yu, and Kilian
  Weinberger.
\newblock Simplifying graph convolutional networks.
\newblock In \emph{International conference on machine learning}, pages
  6861--6871. Pmlr, 2019.

\end{thebibliography}
\end{document}